%% file: neurips_2024.tex
\documentclass{article}

\usepackage[preprint]{neurips_2024}

\usepackage[utf8]{inputenc} 
\usepackage[T1]{fontenc}    
\usepackage{hyperref}       
\usepackage{url}            
\usepackage{booktabs}       
\usepackage{amsfonts}       
\usepackage{nicefrac}       
\usepackage{microtype}      

\usepackage{graphicx}
\usepackage{amsmath}
\usepackage{amssymb}
\usepackage{booktabs}

\usepackage{multirow}
\usepackage{color, colortbl}
\usepackage{subcaption}
\usepackage{tabu}
\usepackage{pifont}
\usepackage{makecell}
\usepackage{paralist}
\usepackage{threeparttable}
\usepackage{enumitem}
\usepackage{xcolor}

\newcommand{\xhdr}[1]{\vspace{4pt} \noindent {\textbf{#1}}}
\newcommand*{\mybox}[2]{\colorbox{#1!30}{\parbox{0.98\linewidth}{#2}}}

\usepackage[capitalize]{cleveref}
\crefname{section}{Sec.}{Secs.}
\Crefname{section}{Section}{Sections}
\Crefname{table}{Table}{Tables}
\crefname{table}{Tab.}{Tabs.}

\newcommand{\gb}[1]{\textcolor{purple}{[{\bf GB:} #1]}}

\newcommand{\xizi}[1]{\textcolor{bittersweet}{[{\bf Xizi:} #1]}}

\newcommand{\mbc}[1]{\textcolor{orange}{[{\bf Mohit:} #1]}}
\newcommand{\done}[1]{{}}

\def\ModelName{\textsc{TimeRefine}}

\title{\ModelName: Temporal Grounding with Time Refining Video LLM}

\author{
  Xizi Wang\\
  Indiana University\\
  \texttt{xiziwang@iu.edu} \\
  \And
  Feng Cheng \\
  UNC Chapel Hill \\
  \texttt{fengchan@cs.unc.edu} \\
  \And 
  Ziyang Wang \\
  UNC Chapel Hill \\
   \texttt{ziyangw@cs.unc.edu} \\
   \AND
   Huiyu Wang \\
   Meta \\
   \texttt{huiyuw@meta.com} \\
   \And
   Md Mohaiminul Islam \\
   UNC Chapel Hill \\
   \texttt{mmiemon@cs.unc.edu} \\
   \And
   Lorenzo Torresani \\
   Meta \\
   \texttt{torresani@meta.com} \\
   \AND
   Mohit Bansal \\
   UNC Chapel Hill \\
   \texttt{mbansal@cs.unc.edu} \\
   \And
   Gedas Bertasius\thanks{co-lead the project} \\
   UNC Chapel Hill \\
   \texttt{gedas@cs.unc.edu} \\
   \And
   David Crandall\footnotemark[1] \\
   Indiana University\\
  \texttt{djcran@iu.edu} \\
}

\begin{document}

\maketitle

\input{sec/0_abstract}    
\input{sec/1_intro}
\input{sec/2_related_work}
\input{sec/3_method}
\input{sec/4_experiment}

\input{sec/5_conclusion}

\small
\bibliographystyle{plainnat}
\bibliography{neurips_2024.bib}

\end{document}

%% file: sec/0_abstract.tex
\begin{abstract}
Video temporal grounding  aims to localize relevant temporal boundaries in a video given a textual prompt. Recent work has focused on enabling Video LLMs to perform video temporal grounding via next-token prediction of temporal timestamps. However, accurately localizing timestamps in videos remains challenging for Video LLMs when relying solely on temporal token prediction.
Our proposed \ModelName{} addresses this challenge in two ways. First, instead of directly predicting the start and end timestamps, we reformulate the temporal grounding task as a temporal refining task: the model first makes rough predictions and then refines them by predicting offsets to the target segment. This refining process is repeated multiple times, through which the model progressively self-improves its temporal localization 
accuracy. Second, to enhance the model's temporal perception capabilities, we incorporate an auxiliary prediction head 
that penalizes the model more if a predicted segment deviates further from the ground truth, thus encouraging the model to make closer and more accurate predictions. 
Our plug-and-play method can be integrated into most LLM-based temporal grounding approaches.
The experimental results demonstrate that \ModelName~achieves 3.6\% and 5.0\% mIoU improvements on the ActivityNet and Charades-STA datasets, respectively.
Code and pretrained models are available at \url{https://github.com/SJTUwxz/TimeRefine}.

\end{abstract}

%% file: sec/1_intro.tex
\section{Introduction}
\label{sec:intro}

\begin{figure}
\centering
\includegraphics[width=0.8\textwidth]{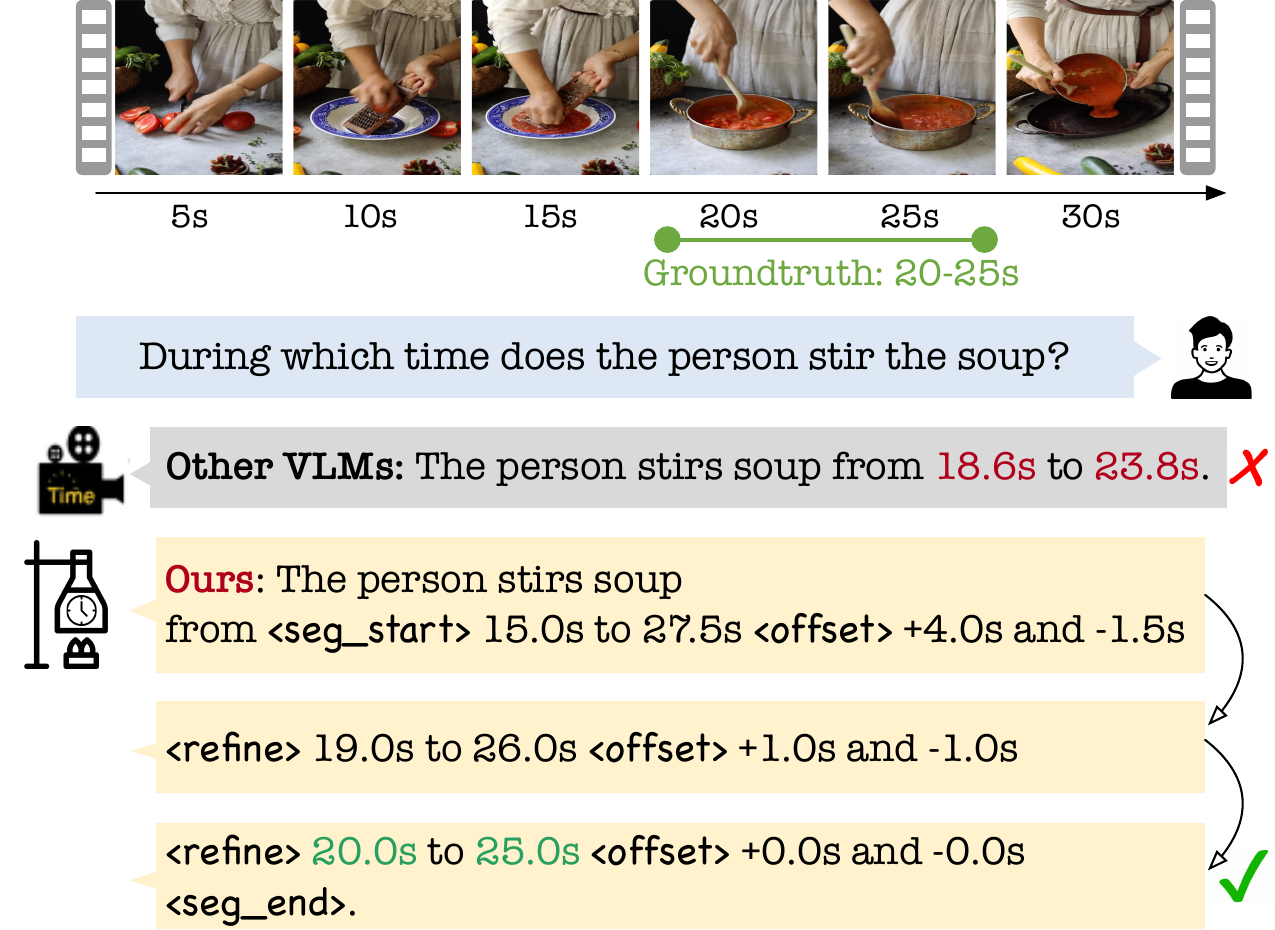}
\caption{Given a text query, existing Video LLMs directly predict the start and end timestamps, which often leads to imprecise localization results. Our model generates a coarse prediction initially and then progressively refines it via temporal offset prediction for more precise temporal localization.
\done{\gb{The question is formulated weirdly. Maybe say, During which time ...? Shouldn't ground truth be two words? Would it be possible to format your predictions in a nicer way? It's a bit difficult to parse. For instance, maybe every new refinement could be shown in a new row or something. Currently it looks a bit messy. I also feel like fully capitalized tags look a bit weird, but maybe it's not a big deal. Maybe you could just use a different font for that.}
\mbc{Can we show the refinement a bit more clearly/cleanly like a multi step process as opposed to one big cluttered paragraph?}
\xizi{edited}}
\vspace{-0.5cm}
}
\label{fig:teaser}
\end{figure}

Video Temporal Grounding (VTG)~\cite{yuan2019semantic,zhang2019man,yuan2019find,zhang2020span,ghosh2019excl,cheng2022tallformer,univtg,yu2019activitynet,zhou2018towards,lei2021detecting,zhang2020does,zheng2025training,chen2020look,chen2018temporally,liu2022unsupervised} is a foundational task in video understanding that aims to localize relevant temporal boundaries in a video given a textual prompt (\textit{e.g.,} ``during which time does the person stir the soup?").
VTG has numerous real-world applications, including anomaly detection~\cite{yao2020and}, sports analytics, security and surveillance, consumer video retrieval, and education.

\done{\mbc{the para structure of the intro needs some changes; currently too many small small 1-2 sentence paras}\xizi{edited}}
Large Language Models (LLMs)~\cite{llama,zeng2022glm,brown2020language,raffel2020exploring,wei2021finetuned,chowdhery2023palm,li2024llava} have demonstrated remarkable capabilities in text understanding and generation in a zero-shot manner. Many recent methods have focused on extending LLMs from language to video. Video LLMs~\cite{videollava,liu2024llavanext,qian2024momentoradvancingvideolarge,timechat,guo2024vtg,videochat,maaz2024videochatgptdetailedvideounderstanding,chen2024sharegpt4video,huang2024vtimellm,cheng2023vindlu,li2023unmasked} have been applied in a wide variety of video understanding tasks, including video captioning~\cite{sun2019videobert,wang2021end,zhao2017hierarchical,pan2020spatio,krishna2017dense,song2015tvsum,li2024mvbench} and video question answering~\cite{yu2019activitynet,xiao2021next,li2024mvbench,wu2024star,xu2016msr,lei2019tvqa+,wang2024loconet}. However, the ability of Video LLMs to perform temporal grounding remains limited, as most pretraining data emphasize detailed content (\textit{i.e.,} ``what") 
understanding rather than temporal perception (\textit{i.e.,} ``when")~\cite{huang2024vtimellm}.

To enhance the temporal perception capabilities of Video LLMs, most  existing work~\cite{guo2024vtg,huang2024vtimellm,timechat} focuses on building high-quality instructional tuning datasets or making architectural changes to better integrate visual features into LLMs  and improve the encoding of temporal tokens. However, the temporal grounding task, which involves regressing timestamps, deviates from the traditional text prediction task used in LLMs. As a result, merely improving the model architecture is insufficient to integrate this task into the LLM framework effectively.

In this work, instead of altering the model architecture, 
we introduce a new perspective to enhance the ability of Video LLMs to handle the temporal grounding task by modifying its learning objective. 
As illustrated in Fig.~\ref{fig:teaser}, unlike other Video LLMs that directly predict start and end timestamps (\textit{e.g.,} ``18.6s to 23.8s") for a given query, we reformulate temporal grounding as a progressive temporal refinement task. Our model initially predicts rough timestamps (\textit{e.g.,} ``15.0s to 27.5s") and then refines its predictions by predicting the offsets (\textit{e.g.,} ``+4.0s and -1.5s") from these initial predictions to reach the final segment prediction. 
We apply multiple such refinement steps to encourage the model to self-correct errors from its previous predictions.

In addition to the time refinement process, we find that applying standard cross-entropy (CE) loss~\cite{goodfellow2016deep} to predict timestamps is suboptimal.
For instance, for a ground truth timestamp of ``20s," CE loss assigns the same penalty for predictions of ``21s" and ``100s," even though the latter is far less accurate. However, for effective temporal boundary learning, the model should be penalized more if the prediction is further away from the ground truth. To address this issue, we supplement the next-token prediction head with an auxiliary prediction head that uses the L1 loss, which penalizes predictions in proportion to their distance from the ground truth. This helps the model learn that predictions closer to the ground truth are preferable, encouraging more accurate temporal localization. 

Our proposed \ModelName~is easily integrable and can be applied to any LLM-based temporal grounding methods. 
Our experimental results show that integrating our method with VTimeLLM~\cite{huang2024vtimellm} leads to 3.6\% and 5.0\% mIoU improvements on the ActivityNet Captions~\cite{krishna2017dense} and Charades-STA datasets~\cite{gao2017tall}, respectively. When applied to VTG-LLM~\cite{guo2024vtg}, we achieve 1.2\% mIoU improvement on the Charades-STA dataset.

%% file: sec/2_related_work.tex
\section{Related Work}
\label{sec:related_work}
\xhdr{Video Large Language Models.}
With the advancement of image-based Large Language Models (LLMs) \citep{blip2, liu2024llavanext}, recent research \citep{videollama, internvideo2, videochat, videollava, zhang2023simple, wang2024videotree} has focused on enhancing the video understanding capabilities of LLMs. Existing models have achieved significant success in various video understanding tasks, such as video question-answering and video captioning. However, they often face challenges with more detailed temporal understanding \citep{liu2024tempcompassvideollmsreally}. 
Most existing methods focuses on building high-quality instructional tuning datasets or making architectural changes, instead, \ModelName{} focuses on modifying the learning objectives to better accommodate the temporal grounding task within the Video LLM framework.

\xhdr{Video Temporal Grounding. }
Video Temporal Grounding (VTG) is crucial to video understanding, as it aims to accurately identify event timestamps within a given video \citep{lin2023univtg}. Traditional task-specific models have been developed to address this task \citep{momentdetr, cgdetr, qddetr, unimd, zala2023hierarchicalvideomomentretrievalstepcaptioning}, typically framing it as a timestamp regression task based on video inputs and user queries. Despite their early success, these models struggle in zero-shot settings, are limited to handling a single task per model, and often require additional fine-tuning for various downstream tasks and datasets. Recent methods \cite{huang2024vtimellm, hawkeye, groundinggpt, lita, zeng2024timesuiteimprovingmllmslong, guo2024tracetemporalgroundingvideo, wang2024groundedvideollmsharpeningfinegrainedtemporal, qu2024chatvtgvideotemporalgrounding, qian2024momentoradvancingvideolarge} leverage the capabilities of powerful Video Large Language Models to perform LLM-based temporal grounding. For instance, VTimeLLM \citep{huang2024vtimellm} directly fine-tunes Video LLMs on VTG datasets. LITA \citep{lita} incorporates SlowFast visual tokens and time tokens into LLM tokenizers to enhance temporal localization capabilities. Momentor \citep{qian2024momentoradvancingvideolarge} addresses time token quantization errors by implementing a time encoder, thus improving fine-grained temporal reasoning. VTG-LLM \citep{guo2024vtg} integrates specialized time tokens and temporal position embeddings to enhance video LLMs' understanding of timestamps.
However, these approaches mainly emphasize the development of mechanisms for encoding visual or temporal features and data collection, often neglecting the fundamental challenges of temporal grounding for Video LLMs. In contrast, \ModelName{} employs a step-by-step refinement schema to enhance the temporal understanding ability, providing a more effective training objective for Video LLMs.

\xhdr{Refinement-based Learning.} The concept of refinement -- initially making a rough prediction and subsequently refining it for greater accuracy -- has been extensively explored across various computer vision domains. For instance, two-stage object detectors~\cite{renNIPS15fasterrcnn,he2017mask,girshick2014rich} utilize a region proposal network to generate candidate bounding boxes, which are then refined by another network to produce final predictions. Similarly, Iterative Error Feedback is commonly applied in tasks that require precise, step-by-step refinement, such as human pose estimation~\cite{carreira2016human}. 
Diffusion models~\cite{ho2020denoising,dhariwal2021diffusion,cheng20254diff} also adopt a step-by-step denoising approach. 
Inspired by the previous works, \ModelName{} proposes a refinement paradigm for the temporal grounding task, which iteratively refines its predictions through next-token prediction. 
To our knowledge, \ModelName{} is the first work to explore the refinement paradigm for LLM-based temporal localization.

%% file: sec/3_method.tex
\section{Method}
\label{sec:method}

\begin{figure*}[t] 
\centering
\includegraphics[width=\textwidth]{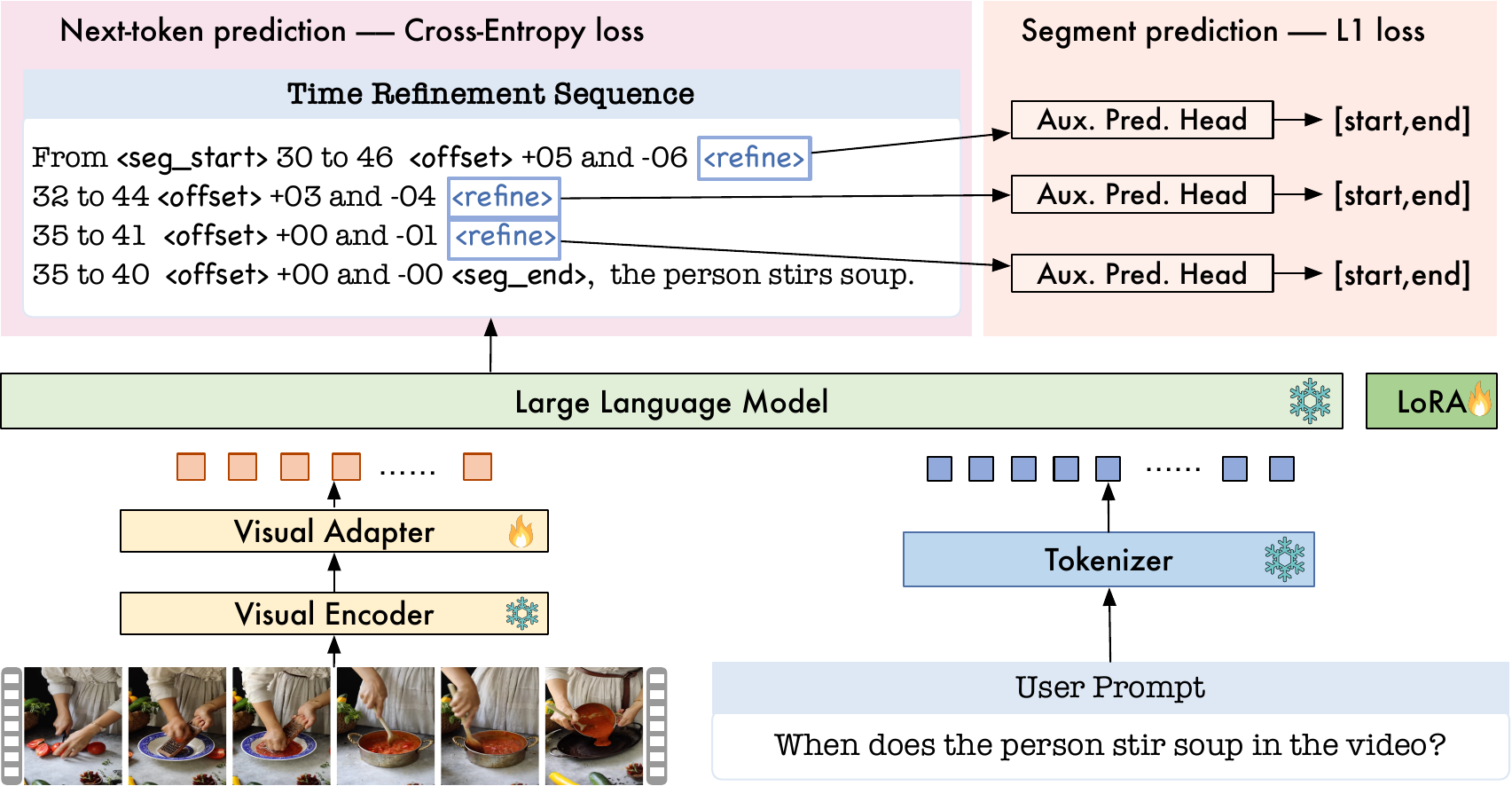}
\caption{
\textbf{Overview of the~\ModelName.} Given a video and a textual user prompt, our model predicts an iterative time refinement sequence, i.e., an initial rough estimation of the boundary, followed by new predictions and offsets based on its previous predictions. The new predictions and offsets can help the model learn how to refine its predictions and correct its errors. Our empirical experiments show that such an iterative temporal refinement strategy can help enhance the temporal perception ability of Video LLMs.
We also complement the Cross-Entropy-based next-token prediction head with an auxiliary prediction head using an L1 regression loss, which encourages the model to learn that closer predictions are preferable. The final prediction is derived from the last predicted segment and its offsets. 
\vspace{-0.3cm}
}
\label{fig:method}
\end{figure*}

\subsection{Problem Formulation}
Given a video $V \in \mathbb{R}^{T \times H \times W \times C}$ and a textual query $Q$ (\textit{e.g.,} ``When does the man have breakfast in the video?"), the model is required to generate a corresponding answer $A$ with localized timestamps (\textit{e.g.,} ``20s to 90s").

Existing approaches~\cite{huang2024vtimellm,guo2024vtg,qian2024momentoradvancingvideolarge,guo2024tracetemporalgroundingvideo,timechat} incorporate video temporal grounding tasks within the Video LLM framework using a next-token prediction scheme, the log-likelihood of the predicted answer conditioned on the textual query and video input could be formulated as:
\begin{equation}
l = \sum_{i=1}^{m} log p(a_i | H_v, H_q, a_{j < i})
\end{equation}

where $H_v = \phi(V)$ represents the video tokens encoded by a pre-trained video encoder $\phi$ (\textit{e.g.,} CLIP~\cite{radford2021learning}), $H_q = \{q_1, q_2, \ldots, q_n\}$ are the tokenized elements of the query, and $H_a = \{a_1, a_2, \ldots, a_m\}$ are the tokenized elements of the answer. The encoded tokens of timestamps (\textit{e.g.,} ``20s") are denoted as temporal tokens, which can be tokenized as standard text tokens (\textit{e.g.,} VTimeLLM~\cite{huang2024vtimellm}) or some special tokens (\textit{e.g.,} VTG-LLM~\cite{guo2024vtg} uses \textless{}TIME-TWO\textgreater).

We identify two key issues in this standard Video LLM-based temporal localization framework:
\begin{itemize}
    \item Directly predicting temporal timestamps in a long video is challenging. Additionally, treating the timestamp prediction task as a token prediction task within the LLM framework results in sparse supervision, leading to inferior temporal localization accuracy. The model receives supervision signals from only two temporal tokens (e.g., a start and end token for each segment) while processing hundreds of visual and textual tokens. In comparison, traditional temporal grounding models (\textit{e.g.,} ActionFormer~\cite{zhang2022actionformer}, TallFormer~\cite{cheng2022tallformer}) regress hundreds of bounding boxes for each input, providing much stronger supervision signals. 
    \item The Cross-Entropy (CE) loss used in next-token prediction is not well-suited for temporal prediction tasks. For instance, for a timestamp ``20s", CE assigns the same penalty for predictions of ``21s" and ``100s", despite the latter being significantly more erroneous. This issue prevents the model from learning that closer predictions are preferable in temporal tasks.
\end{itemize}

To address these issues, we propose \ModelName, which is detailed in the following section.

\subsection{\ModelName}
To address abovementioned issues, we introduce \ModelName~, which proposes a step-by-step refinement paradigm to the LLM-based temporal grounding task. This iterative refinement strategy helps the model learn to self-correct its errors in boundary perception, thus enhancing the Video LLM's temporal understanding ability.

First, we reformulate the temporal token prediction task into an iterative time refinement task. Given a video and a user prompt, the model is required to make multiple rounds of predictions. In each round, it makes an initial rough prediction and then learns to self-correct the errors of this rough prediction. This scheme not only provides a stronger supervision signal but also transforms the problem from directly predicting precise timestamps to making coarse-to-fine predictions, akin to how humans localize moments in videos. Second, we introduce an auxiliary time prediction head to enhance the Video LLM's temporal perception capability. This auxiliary head is optimized using L1 loss, which helps the model learn that closer predictions are preferable in temporal grounding tasks.

\subsubsection{Architecture}

Our method is agnostic to the exact model architecture and can be applied to any LLM-based VTG method that follows the formulation described above. For clarity, we visualize a common architecture of Video LLMs in Fig.~\ref{fig:method}. This architecture consists of a visual encoder, a visual adapter, a text tokenizer, and a large language model (LLM). The visual encoder is typically a CLIP-like model~\cite{radford2021learning} used to extract visual features. The visual adapter projects these visual features into the language space using MLPs~\cite{rumelhart1986learning} or Q-Formers~\cite{blip2}. The LLM takes the visual embeddings of the video and the embedded text tokens of user prompts as input to output the answers. We retain the core architecture unchanged but modify the learning objectives to include 1) iterative refinement of temporal segments and 2) an auxiliary segment prediction head, which we discuss below.

\subsubsection{Iterative Time Refinement}

We modify the direct timestamp prediction scheme in two ways. First, instead of predicting the timestamp directly, which is challenging for Video LLMs, we employ a coarse-to-fine approach. The model first makes a rough prediction and then predicts the offsets to the target segment, encouraging the model to learn to self-correct its errors. Second, making accurate predictions in a single step is difficult, so we implement the coarse-to-fine prediction scheme iteratively. This allows the model multiple opportunities to correct its errors, thereby enhancing its temporal localization accuracy.

Specifically, for the start and end timestamp in a segment $S = (s, e)$ that the model needs to predict, we convert it into an iterative time refinement prediction sequence: 
\begin{equation}
S' = ((s_0, e_0, o_0^s, o_0^e), \ldots, (s_K, e_K, o_K^s, o_K^e))
\end{equation}
where $(s_0, e_0, o_0^s, o_0^e)$ represents a single coarse-to-fine prediction step, $s = s_k + o_k^s$, $e_i = e_k + o_k^e$, and $o_k^s$ and $o_k^e$ are offsets to the target segment. $K$ denotes the number of refinement steps.
During training, we format $S_i'$ into a sequence with additional control tokens:
\noindent \mybox{gray}{$<$seg\_start$>$ \\ 
$s_0$ to $e_0$ $<$offset$>$ $o_0^s$ and $o_0^e$ \\ 
\textbf{$<$refine$>$} $s_1$ to $e_1$ $<$offset$>$ $o_1^s$ and $o_1^e$ \\
... \\
\textbf{$<$refine$>$} $s_K$ to $e_K$ $<$offset$>$ $o_K^s$ and $o_K^e$ \\
$<$seg\_end$>$}

Our iterative time refinement scheme requires the model to learn three tasks: (1) make rough predictions ($s_k$ and $e_k$) based on user inputs, (2) self-correct the error it makes by predicting $o_k^s$ and $o_k^e$), and (3) initiate new predictions and refinements based on its previous refinement steps. Our scheme provides much stronger supervision signals compared to the direct timestamp prediction used in existing Video LLMs. It also enables the model to learn self-correction, which can further enhance temporal prediction accuracy.

\xhdr{Training Sequence Generation.} A crucial aspect of our reformulated task is the generation of the training sequence. We draw inspiration from diffusion models~\cite{ho2020denoising,dhariwal2021diffusion} that progressively add less noise during the denoising process to approximate the ground truth. Specifically, we define $K$ Gaussian distributions $\{\mathcal{N}_k(0, \sigma_k^2), k=1, \ldots, K\}$, where $\sigma_{k}$ and $K$ are predefined parameters. Given a ground truth segment $S_i = (s_i, e_i)$, we sample the offsets $o_k^s$ and $o_k^e$ from these distributions:
\begin{align}
    o_k^s &\sim \mathcal{N}_k(0, \sigma_k^2), \quad o_k^e \sim \mathcal{N}_k(0, \sigma_k^2) \\
    s_k &= s - o_k^s, \quad e_k = e - o_k^e
\end{align}
We resample at refinement step $k$ if $s_{k}$ or $e_{k}$ are less than 0 or greater than the video duration.

\subsubsection{Temporal Perception}
\label{sec:temporal perception}
The Cross-Entropy (CE) loss~\cite{goodfellow2016deep}, commonly used in LLM-based video temporal grounding methods, is less suitable for temporal timestamp predictions, which are continuous variables. For instance, for a ground truth timestamp at $20s$, CE loss assigns the same penalties for the following two predictions: (i) $p(t = 20s) = 0.1, p(t = 21s) = 0.9$ and (ii) $p(t = 20s) = 0.1, p(t = 100s) = 0.9$. This approach is counterintuitive, as the latter prediction is significantly more inaccurate.

To address this issue, we propose adding an auxiliary prediction head attached to the $<$refine$>$ token to predict the segment start and end:
\begin{equation}
    \hat{S} = \text{Linear}(h[\textless\text{refine}\textgreater])
\end{equation}
where $h[\textless\text{refine}\textgreater]$ is the embedding of the \textless\text{refine}\textgreater{} token from the model's last layer. The output $\hat{S}$ is supervised by the L1 loss between predictions and ground truth: $L1 = |\hat{S} - S|$. Unlike Cross-Entropy loss, L1 loss penalizes predictions more heavily when they are further from ground truth, thus enhancing the model's temporal perception ability. To maintain the integrity of the Video LLM, we retain the original Cross-Entropy loss and add this L1 loss as an auxiliary supervision signal. We experimented with other loss functions, such as GIoU~\cite{rezatofighi2019generalized} loss, but did not observe improvements. See Sec.~\ref{sec:ablate_temporal_perception} for more design choices.

\subsection{Training and Inference}

\xhdr{Training.} The total loss is the sum of the original Cross-Entropy loss and the L1 loss of the auxiliary linear branch, as defined in Sec.~\ref{sec:temporal perception}.
\begin{equation}
    L = \frac{1}{m}\sum_{i=1}^{m}\text{cross\_entropy}(\hat{a}_i, a_i) + \lambda\frac{1}{2|S|}|\hat{S} - S|
\end{equation}
where $\hat{a}_i$ and $a_i$ are the predicted and ground-truth answer tokens (including temporal tokens), $|S|$ is the number of segments in the answer, and $\lambda$ is a hyperparameter to balance these two losses.

\xhdr{Inference.} Our answer generation process follows the next-token prediction scheme in Video LLMs. The auxiliary prediction head (Sec.~\ref{sec:temporal perception}) is discarded during inference\. For the prediction of the temporal segment $S$, we use the predictions from the final refinement step, $s_K + o_K^s$ and $e_K + o_K^e$, as the predicted start and end timestamps.

\subsection{Implementation Details}
The proposed~\ModelName~is architecture-agnostic and can be applied to most LLM-based VTG methods. We first introduce the hyperparameters of our \ModelName, and then explain how we apply \ModelName{} to other VTG methods (\textit{i.e.,}  VTimeLLM~\cite{huang2024vtimellm} and VTG-LLM~\cite{guo2024vtg}).

\subsubsection{\ModelName}

\xhdr{Iterative Time Refinement Scheme.} We reformat the training data by replacing the direct start and end timestamps in grounding-related QA pairs with time-refinement sequences, while leaving non-grounding QA pairs unchanged. To construct the iterative time refinement sequence, we choose the number of refinement steps of each time segment as $K=4$. The $K$ Gaussian distributions have fixed standard deviations sorted in decreasing order: $\sigma^2=\{5,3,1,0\}$ (in seconds). Such a design will progressively add less noise in the refinement process to approximate the ground truth, which can simulate a coarse-to-fine refinement process.

\xhdr{Temporal Perception.} In addition to the Cross-Entropy loss, we add the L1 loss with a weight of $\lambda=10$ to balance the losses. We show ablations of loss choices in Sec.~\ref{sec:ablate_temporal_perception}.

\subsubsection{Integrating with other VTG methods}

\xhdr{VTimeLLM.} To enable direct comparison with VTimeLLM~\cite{huang2024vtimellm}, we use the same training settings and data as VTimeLLM. We initialize~\ModelName~from VTimeLLM's stage one checkpoint so that visual features are already aligned with LLM's semantic space. We then train our model on the stage two and stage three training data for 4000 steps and 1000 steps, respectively, with a learning rate of $1\times10^{-4}$ on a single 8-A5000 (24G) machine. The LoRA rank is set to 64 and alpha to 128. 

\xhdr{VTG-LLM.} We also apply \ModelName{} on VTG-LLM, and train the model on the same training data of VTG-LLM, i.e., VTG-IT-120k~\cite{guo2024vtg} and a randomly sampled subset (97k) from the Valley dataset~\cite{luo2023valley}. All the hyperparameters (batch size, learning rate, optimizers, etc.) are kept the same as the original VTG-LLM.

%% file: sec/4_experiment.tex
\begin{table*}
    \centering
    \footnotesize
    \setlength{\tabcolsep}{1pt}
    \resizebox{\textwidth}{!}{\begin{tabu*}{lccccccccccc}
        \toprule
        \multirow{3}{*}{\bf Method} & \multicolumn{8}{c}{\bf Temporal Grounding} & \multicolumn{3}{c}{\bf Dense Captioning} \\
        \cmidrule{2-9} \cmidrule{10-12}
         & \multicolumn{4}{c}{\bf ActivityNet Captions} & \multicolumn{4}{c}{\bf Charades-STA} & \multicolumn{3}{c}{\bf ActivityNet Captions}\\
        \cmidrule(lr){2-5} \cmidrule(lr){6-9} \cmidrule(lr){10-12}
        & R@0.3 & R@0.5 & R@0.7 & mIoU & R@0.3 & R@0.5 & R@0.7 & mIoU & SODA\_c & CIDEr & METEOR\\
        \midrule
        \multicolumn{6}{l}{\textit{Traditional Video LLMs}} \\
        VideoChat~\cite{videochat} & 8.8 & 3.7 & 1.5 & 7.2 & 9.0 & 3.3 & 1.3 & 6.5 & 0.9 & 2.2 & 0.9 \\
        VideoLLaMA~\cite{videollama} & 6.9 & 2.1 & 0.8 & 6.5 & 10.4 & 3.8 & 0.9 & 7.1 & 1.9 & 5.8 & 1.9 \\
        Video-ChatGPT~\cite{maaz2024videochatgptdetailedvideounderstanding} & 26.4 & 13.6 & 6.1 & 18.9 & 20.0 & 7.7 & 1.7 & 13.7 & 1.9 & 5.8 & 2.1 \\
        Valley~\cite{luo2023valley} & 30.6 & 13.7 & 8.1 & 21.9 & 28.4 & 1.8 & 0.3 & 21.4 & 0.3 & 1.8 & 0.8 \\
        \midrule
        \multicolumn{6}{l}{\textit{Temporal Grounding Video LLMs}} \\
        GroundingGPT~\cite{groundinggpt} &- &- &- &- & - & 29.6 & 11.9 & - & - & - & -\\
        Momentor~\cite{qian2024momentoradvancingvideolarge} & 42.9 & 23.0 & 12.4&  29.3& 42.6& 26.6& 11.6 &28.5 & 2.3& 14.9& 4.7 \\
        TimeChat~\cite{timechat} &- &- &- &- & - & 32.2 & 13.4 & - & - & - & -\\
        HawkEye~\cite{hawkeye} & \textbf{49.1} & 29.3 & 10.7 &32.7& 50.6&31.4&14.5 & 33.7 & \\
        \midrule 
        VTG-LLM~\cite{guo2024vtg} & - & - & - & - & 51.2 & 33.8 & 15.7 & 34.4 & - & - & -\\
        VTG-LLM-\ModelName & -&- &- &- & 55.8 & 35.1 & 16.7 & 35.6 & - & - & - \\
        \midrule
        VTimeLLM-7B~\cite{huang2024vtimellm} & 44.0 & 27.8 & 14.3 & 30.4 & 51.0 & 27.5 & 11.4 & 31.2 & 5.8 & 27.6 & 6.8 \\
        VTimeLLM-\ModelName(7B) & 48.0 & \textbf{33.6} & \textbf{17.5} & \textbf{34.0} & \textbf{56.9} & \textbf{38.6} &	\textbf{16.4} &	\textbf{36.2} & \textbf{6.1} &	\textbf{28.6}&\textbf{7.0}\\
        \bottomrule
    \end{tabu*}}
    \vspace{-0.1cm}
    \caption{\textbf{Comparison with state-of-the-art Video LLMs in temporal grounding and dense video captioning on ActivityNet Captions and Charades-STA.} We observe that~\ModelName~achieves the best performance on the two datasets on most metrics. This demonstrates that iterative time refinement with an added auxiliary prediction head can help temporal understanding and prediction in temporal grounding-related tasks.}
    \vspace{-0.3cm}
    
    \label{tab:sota_compare}
\end{table*}

\section{Experiments}
\label{sec:experiment}

\subsection{Evaluation Setups}

\xhdr{Evaluation Datasets.} Following prior work~\cite{huang2024vtimellm,guo2024vtg}, we employ ActivityNet Captions~\cite{krishna2017dense} and Charades-STA dataset~\cite{gao2017tall} as evaluation datasets. We measure the Intersection over Union (IoU)~\cite{everingham2010pascal} between the predicted time segments and the ground truth time segments. We report the Recall@1 at IoU thresholds of 0.3, 0.5 and 0.7, as well as mean IoU (mIoU). 
Though not the primary focus of our work, we also evaluate our method on the dense captioning task using the ActivityNet Captions dataset. For evaluation, we use SODA\_c~\cite{fujita2020soda} to assess storyline consistency, CIDEr~\cite{vedantam2015cider} to evaluate caption quality, and METEOR~\cite{banerjee2005meteor} to measure temporal localization accuracy, averaged across different IoU thresholds.

\xhdr{Baselines.} Besides the baseline methods VTimeLLM and VTG-LLM, we further compare the proposed method with two types of Video LLMs: 1) traditional Video LLMs like VideoChat-7B~\cite{videochat}, VideoLLaMA-7B~\cite{videollama} and VideoChatGPT-7B~\cite{maaz2024videochatgptdetailedvideounderstanding}, Valley-7B~\cite{luo2023valley} and 2) Video LLMs that are specifically designed for VTG: GroundingGPT~\cite{groundinggpt}, TimeChat~\cite{timechat}, Momentor~\cite{qian2024momentoradvancingvideolarge} and HawkEye~\cite{hawkeye}.

\subsection{Main Results}
In Table~\ref{tab:sota_compare}, we apply the proposed~\ModelName~on VTimeLLM~\cite{huang2024vtimellm} and evaluate the performance on ActivityNet Captions~\cite{krishna2017dense} and Charades-STA~\cite{gao2017tall}. With the same training settings and data, it surpasses VTimeLLM-7B by \textbf{3.2\%} in R@0.7 and \textbf{3.6\%} in mIoU on the ActivityNet Captions dataset. Though not our purpose, we can also observe a small performance increase on the dense video captioning task, showing that the proposed method not only strengthens temporal grounding ability but also benefits other grounding-related tasks. 

We further compare the results with existing Video LLMs. It is important to note that different Video LLMs are trained on varying datasets, which complicates fair comparisons. Nonetheless, \ModelName~achieves the best performance on most metrics among all 7B Video LLMs on both the ActivityNet Captions and Charades-STA datasets.

To demonstrate the plug-and-play adaptability of the proposed \ModelName~with various Video LLMs, we further evaluate its application on VTG-LLM~\cite{guo2024vtg} with results in Table~\ref{tab:sota_compare}. For zero-shot temporal grounding on Charades-STA, \ModelName~outperforms VTG-LLM by \textbf{4.6\%} on R@0.3,  \textbf{1.3\%} on R@0.5 and \textbf{1.0\%} on R@0.7. While VTG-LLM improves temporal grounding accuracy by integrating knowledge about timestamps, \ModelName~further boosts this capability through iterative time refinement strategy and auxiliary temporal perception supervision. 

\begin{figure}[t] 
\centering
\includegraphics[width=0.8\textwidth]{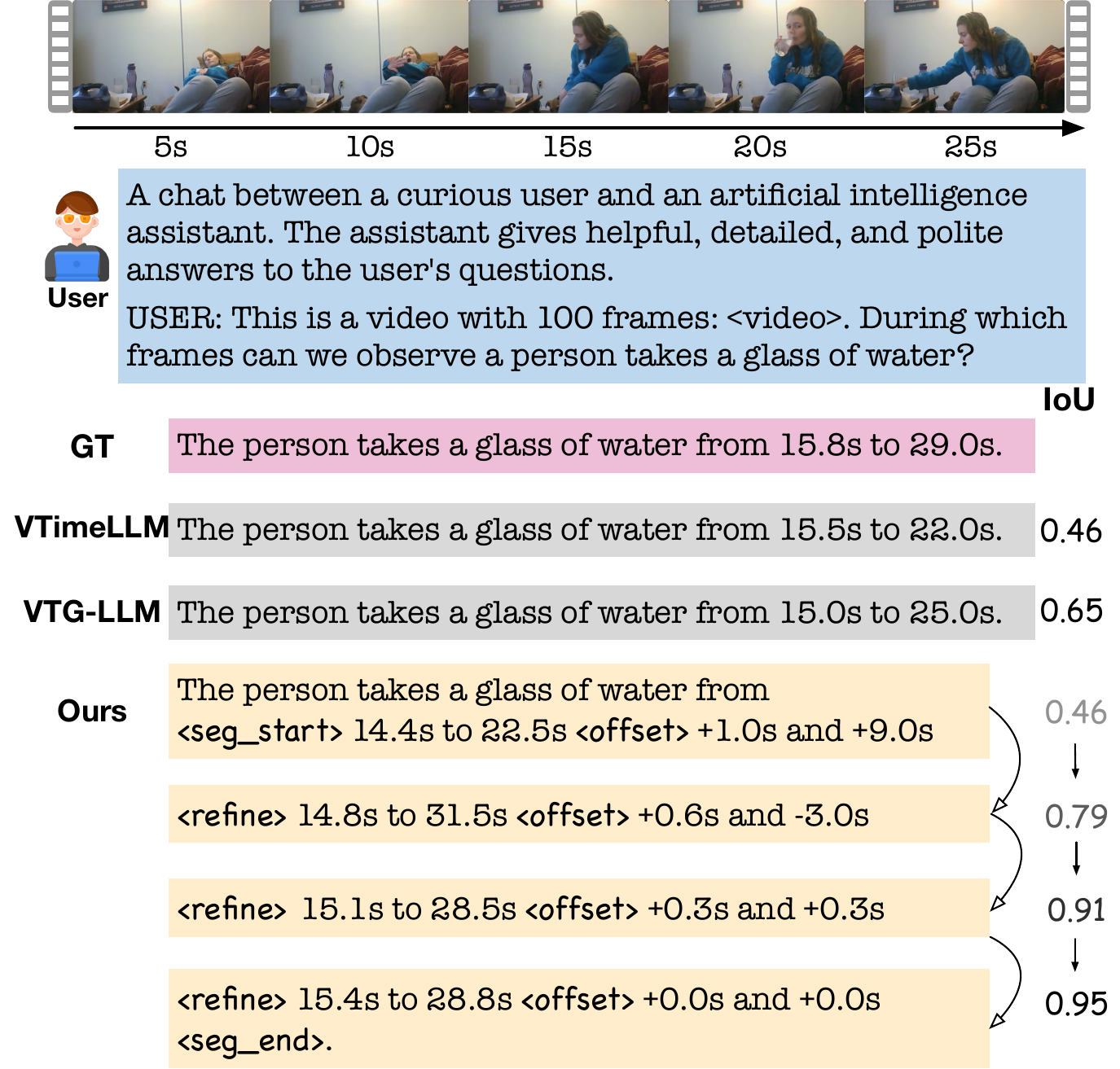}
\caption{\textbf{Zero-shot case study.} We compare the output of VTimeLLM, VTG-LLM and \ModelName~on a video from Charades-STA dataset. Our method iteratively refines the segment predictions. The final prediction achieves an IoU of 0.95, which is the highest among all predictions. 
}
\label{fig:case_study}
\end{figure}

\subsection{Qualitative Evaluation}

In Fig.~\ref{fig:case_study}, we compare the output of VTimeLLM~\cite{huang2024vtimellm}, VTG-LLM~\cite{guo2024vtg} and \ModelName~on the Charades-STA dataset in a zero-shot setting. VTimeLLM and VTG-LLM directly predict one segment, resulting in IoU of $0.46$ and $0.65$. In contrast, VTimeLLM-\ModelName~estimates a rough boundary of the segment at first with $IoU=0.46$.  Then it progressively refines this estimation over $3$ additional refinement steps, providing a new segment and offsets at each refinement step. The final prediction of \ModelName~achieves the highest IoU of $0.95$, showing the effectiveness of iterative time refinement strategy and auxiliary supervision signal.

\subsection{Design Choices}
\label{sec:ablations}

In this section, we examine the design choices of ~\ModelName{}  and provide detailed ablations about the iterative time refinement strategy and the auxiliary prediction head for temporal perception. We begin by analyzing the refinement task design, specifically, the refinement goal. Next, we explore the multi-step refinement sequence generation strategy, such as the number of refinement steps and the appropriate amount of noise to add. Finally, we investigate the use of auxiliary prediction branches to enhance the model's temporal perception capabilities and the method for decoding target timestamps from our predictions. Unless otherwise specified, all results are based on the evaluation of \ModelName{} on VTimeLLM-7B \cite{huang2024vtimellm} for the temporal grounding task of ActivityNet Captions.

\subsubsection{Time Refinement Task Design}

\begin{table}
    \centering
    \small
    \setlength{\tabcolsep}{4pt}
    \begin{tabu}{lcccc}
        \toprule
        \multirow{2}{*}{\bf Method} & \multicolumn{4}{c}{\bf ActivityNet Captions}\\
        & R@0.3 & R@0.5 & R@0.7 & mIoU \\
        \midrule
       No Refinement & 44 & 27.8 & 14.3 & 30.4 \\
       IoU prediction & 44.8 & 28.3 & 14.8 & 30.9 \\
       offset prediction & \textbf{45.2} & \textbf{29.0} & \textbf{15.4} & \textbf{31.5}\\
       \bottomrule
    \end{tabu}
    \vspace{0.2cm}
    \caption{\textbf{Design choices of the refinement task.} We explore predicting IoU and temporal offsets at each refinement step. Predicting  offsets can better guide the model on how to refine boundaries. }
    
    \label{tab:ablate_refine_task}
\end{table}

The first question is how to design a refinement task that helps the Video LLM 
better learn temporal grounding. In Table~\ref{tab:ablate_refine_task}, we explore three options:
{\setlength{\parindent}{0em}
\begin{itemize}
\vspace{-1mm}
    \item \textbf{No Refinement}: The model directly predicts the start and end timestamps (e.g., ``from 10s to 20s").
    \item \textbf{IoU Prediction}: The model first predicts the timestamps and then predicts the Intersection over Union (IoU) between its prediction and the ground truth (e.g., ``from 10s to 20s, the IoU is 0.6; from 11s to 18s, the IoU is 0.9"). The model can incorporate this feedback for the self-correction process.
    \item \textbf{Offset Prediction}: The model first predicts the timestamps and then predicts the offset of its prediction relative to the target for each timestamp (e.g., ``from 10s to 20s, the offset is +2s and -3s; from 12s to 17s, the offset is 0s and 0s"). This information can benefit the self-correction process by providing additional guidance on how to refine the prediction. 
\end{itemize}
}
We apply Gaussian noise with a standard deviation of 5 seconds to generate the training samples for the IoU prediction and offset prediction. 
From Table~\ref{tab:ablate_refine_task}, we observe that both the IoU and offset prediction methods achieve higher performance compared to the no-refinement variant. This suggests that the model benefits from being required to consider its predictions more thoroughly, akin to the ``chain of thought" concept. Offset predictions achieve the best performance, outperforming the no-refinement approach by \textbf{1.1\%} in mIoU. We conjecture that this is because offset prediction provides more direct and straightforward supervision than IoU prediction.

\textit{We use the offset prediction scheme from now on.}

\begin{table}
    \centering
    \small
    \setlength{\tabcolsep}{4pt}
    \begin{tabu}{lcccc}
        \toprule
        \multirow{2}{*}{\bf Gaussian Std Dev} & \multicolumn{4}{c}{\bf ActivityNet Captions}\\
        & R@0.3 & R@0.5 & R@0.7 & mIoU \\
        \midrule
        \multicolumn{5}{l}{\textit{\# Refinement steps}} \\
       2:\{5,0\}  & 45.2  & 29.0 & 15.4 & 31.5  \\
       4:\{5,3,1,0\}  & \textbf{46.2} & \textbf{30.7} & \textbf{16.3} & \textbf{32.4} \\
       6:\{5,4,2,1,0.5,0\}  & 46.0 & 30.8 & 16.1 & 32.3 \\
       8:\{5,4,3,2,1.5,1,0.5,0\} & 45.8 & 30.4 & 16.1 & 32.0 \\
        \midrule
       \multicolumn{5}{l}{\textit{Fixed Std. \& Vary scale of noise}} \\
       \{2,1.2,0.4,0\} & 45.7 & 30.6 & 16.2 & 32.0 \\ 
       \{5,3,1,0\} & \textbf{46.2} & \textbf{30.7} & \textbf{16.3} & \textbf{32.4}\\
       \{10,6,2,0\} & 45.2 & 29.9 & 15.5 & 31.9 \\
       \{20,12,4,0\} & 45.0  & 29.5 & 15.1 & 31.6\\
       \midrule
       \multicolumn{5}{l}{\textit{Adaptive Std. w.r.t Video Length}} \\
       \{0.2, 0.1, 0.05, 0\} & 45.3 & 30.4 & 15.7 & 31.5 \\
       \{0.1, 0.05, 0.01, 0\} & 45.8  & 30.8 & 16.0 & 31.8\\
        \bottomrule
    \end{tabu}
    \vspace{0.2cm}
    \caption{\textbf{Ablation studies on the number of refinement steps and added noise}. The numbers in ``\{\}" represent the standard deviations of the Gaussian noises added at each step, measured in seconds. In the bottom part (Adaptive std w.r.t. the video length), the standard deviations are expressed as a proportion of the video durations.
    We vary the number of refinement steps from 2 to 8, and observe that the performance saturates at the number of refinement steps set to 4.
    Since we sample noise from Gaussian distribution centering at the target timestamp, we further investigate the effect of fixed standard deviation versus adaptive standard deviation with regard to the video length. Experiments indicate that fixed standard deviation achieves better performance.}
    
    \label{tab:ablate_noise}
\end{table}

\subsubsection{Refinement Sequence Generation Strategy}
\label{sec:refine_seq_gen}

In this section, we investigate how to generate the refinement sequence via $K$ Gaussian distributions. We show the results in Table~\ref{tab:ablate_noise}.

\xhdr{How many refinement steps $K$ does the model need?} In the iterative time refinement process, we investigate how varying the number of refinement steps $K$, from 2 to 8, affects the model's temporal prediction performance. We found out that setting the number to 4 brings the most improvement, with performance gains plateauing beyond this point. This shows that predicting and refining multiple temporal segments is beneficial to temporal grounding task. 

\xhdr{How much noise does the model need?} For the $K=4$ Gaussian distributions with fixed standard deviations, we tune the amount of noise to be added. Starting from standard deviations of $\{5,3,1,0\}$, we adjust the noise level by factors of 0.4, 1, 2 and 4 to try different scale of noise. Experiment results demonstrates that adding smaller amount of noise is more effective. Whereas larger scale of noise can disrupt the model's learning process. 

\xhdr{Does the noise need to be adaptive to the video duration?} 
We validate this by setting the standard deviations to be adaptive to the length of the video, adding stronger noise for longer videos. Instead of using fixed standard deviations, we adopt adaptive standard deviations of $\{0.2,0.1,0.05,0\}$ and $\{0.1,0.05,0.01,0\}$, where each value represents a fraction of the video duration (e.g., $0.2$ corresponds to $0.2\times$ video duration). In this way, we add larger noise to longer videos. The bottom part of Table~\ref{tab:ablate_noise} shows that using adaptive standard deviations results in a \textbf{0.6\%} lower mIoU compared to using fixed standard deviations. This outcome is reasonable, as the temporal dynamics within a video are independent of its length, suggesting that a consistent noise distribution should be applied to both short and long videos.

\textit{From now on, we use $K=4$ refinement steps with std. $\{5,3,1,0\}$.}

\begin{table}
    \centering
    \setlength{\tabcolsep}{4pt}
    \begin{tabu}{lcccc}
        \toprule
        \multirow{2}{*}{\bf Method} & \multicolumn{4}{c}{\bf ActivityNet Captions}\\
        & R@0.3 & R@0.5 & R@0.7 & mIoU \\
        \midrule
       \multicolumn{5}{l}{\textit{Auxiliary Loss Type}} \\
       None &  46.2 & 30.7 & 16.3 & 32.4 \\
       L1         & \textbf{48.0} & \textbf{33.6} & \textbf{17.5} & \textbf{34.0}\\
       L1 + GIoU  & 47.5 & 33.0 & 16.4 & 33.7 \\
       L2         & 46.6 & 32.9 & 15.9 & 33.5 \\
       L2 + GIoU  & 46.8 & 32.7 & 15.3 & 33.4 \\
        \bottomrule
    \end{tabu}
    \vspace{0.2cm}
    \caption{\textbf{Ablation studies on the auxiliary loss type of temporal perception module}. We first explore multiple loss types including L1, L1+GIoU, L2 and L2+GIoU loss. Results show that L1 loss is the most effective, while adding GIoU loss doesn't bring further performance increase. }
    
    \label{tab:ablate_aux}
\end{table}

\begin{table}
    \centering
    \setlength{\tabcolsep}{4pt}
    \begin{tabu}{lcccc}
        \toprule
        \multirow{2}{*}{\bf Method} & \multicolumn{4}{c}{\bf ActivityNet Captions}\\
        & R@0.3 & R@0.5 & R@0.7 & mIoU \\
        \midrule
       \multicolumn{5}{l}{\textit{Decode Segment Predictions}} \\
       seq. prediction first step & 45.7 & 31.3 & 15.3 & 31.5 \\
       seq. prediction last step  & \textbf{48.0} & \textbf{33.6} & \textbf{17.5} & \textbf{34.0} \\
       aux. prediction head & 47.9 & 33.4 & 17.3 & 33.9\\
       aux. \& seq. prediction & 48.0 & 33.5 & 17.5 & 34.0\\
        \bottomrule
    \end{tabu}
    \vspace{0.2cm}
    \caption{\textbf{Ablation studies on timestamp decoding methods}. It can be observed that using the last step of sequence prediction achieves 2.5\% higher mIoU than using the first step. This aligns with the training sequence, where the last step in the refinement process is the closest to the target segment. On the other hand, utilizing predictions from the auxiliary prediction head, the last step of sequence prediction, or merging the two predictions yields similar performance. This indicates that we can discard the auxiliary head after training, making our method compatible with existing token generation frameworks.}
    
    \label{tab:ablate_decode}
\end{table}

\subsubsection{Temporal Perception}
\label{sec:ablate_temporal_perception}

\xhdr{Does an auxiliary prediction branch help temporal perception?} 
In Table~\ref{tab:ablate_aux}, adding an auxiliary branch optimized with L1 loss leads to a \textbf{1.6\%} higher mIoU comparing to no auxiliary loss. This proves that an auxiliary supervision signal can complement the original Cross-Entropy loss and further enhance the model's temporal perception capability. 

\xhdr{Which loss can better help temporal prediction tasks?} We employ L1 loss, L2 loss and GIoU loss as the auxiliary supervision signal in addition to the Cross-Entropy loss, since these losses are commonly used in detection and grounding tasks~\cite{renNIPS15fasterrcnn,redmon2018yolov3,bochkovskiy2020yolov4,rezatofighi2019generalized,tan2020efficientdet,zhu2020deformable}. In Table~\ref{tab:ablate_aux}, experiments show that L1 loss is the most effective loss, while further adding GIoU loss doesn't bring much improvement. 

\subsubsection{Segment Decoding}
\label{sec:decoding}

Each refinement process involves multiple steps, allowing us to select the segment predicted at either the initial or final step. Alternatively, we can use the segment predicted by the auxiliary head or combine it with the sequence-based prediction. Table~\ref{tab:ablate_decode}
 presents the results for these decoding strategies. We can observe that 1) evaluating using segments predicted at the final step significantly outperforms using the segment from the initial step. This is consistent with our sequential refinement design, where the final segment is intended to match the target; and 2) utilizing predictions from either the auxiliary prediction head, last step of sequence prediction or combining the two gives similar performance. This means that the auxiliary head can be discarded after training, making our method compatible with existing token generation frameworks.

%% file: sec/5_conclusion.tex
\section{Conclusion}

In this work, we propose \ModelName{} to enhance the capability of Video LLMs in performing temporal grounding. Unlike previous approaches that focus on data curation or architectural enhancements, we focus on refining the learning objective to better suit temporal grounding within the Video LLM framework. We transform the timestamp prediction task into an iterative error refinement task. Additionally, we complement the next-token prediction with an auxiliary L1 prediction head, encouraging the model to make closer and more accurate predictions. Our method is architecture-agnostic, allowing us to apply it to two recent state-of-the-art LLM-based VTG methods, VTimeLLM and VTG-LLM, demonstrating promising performance improvements with minimal training cost overhead. One limitation of our method is the need to predict several times more temporal-related tokens compared to prior methods. However, this drawback is mitigated when users are interested in both temporal grounding and question-answering tasks, where the model generates significantly more textual tokens compared to temporal tokens. Future research directions include further refinement of task and sequence designs and exploring how temporal grounding can enhance video question-answering tasks.

\section*{Acknowledgement}
This work was supported in part by the National Science Foundation under award DRL-2112635 to the AI Institute for Engaged Learning. Any opinions, findings, and conclusions or recommendations expressed in this material are those of the author(s) and do not necessarily reflect the views of the National Science Foundation. Authors from UNC Chapel Hill were supported by the Sony Faculty Innovation award, Laboratory for Analytic Sciences via NC State University, and ONR Award N00014-23-1-2356.